\documentclass{article}

\usepackage[final]{neurips_2024}

\makeatletter
\renewcommand{\@noticestring}{}
\makeatother

\usepackage{graphicx}
\usepackage[utf8]{inputenc} 
\usepackage[T1]{fontenc}    
\usepackage{hyperref}       
\usepackage{url}            
\usepackage{booktabs}       
\usepackage{amsfonts}       
\usepackage{nicefrac}       
\usepackage{microtype}      
\usepackage{xcolor}         
\usepackage{siunitx}
\usepackage{transparent}
\setcitestyle{numbers}

\title{ALVI Interface: Towards Full Hand Motion Decoding for Amputees Using sEMG}

\author{
    \textbf{Aleksandr Kovalev} \quad
    \textbf{Anna Makarova} \quad
    \textbf{Petr Chizhov} \quad
    \textbf{Matvey Antonov} \\
    \textbf{Gleb Duplin} \quad
    \textbf{Vladislav Lomtev} \quad
    \textbf{Viacheslav Gostevskii} \quad
    \textbf{Vladimir Bessonov} \\
    \textbf{Andrey Tsurkan} \quad
    \textbf{Mikhail Korobok} \quad
    \textbf{Aleksejs Timčenko} \\
    \texttt{ALVI Labs} \\
    \texttt{\href{mailto:koval.alvi@gmail.com}{koval.alvi@gmail.com}}}

\begin{document}
\maketitle
\begin{abstract}

We present a system for decoding hand movements using surface EMG signals. The interface provides real-time (25 Hz) reconstruction of finger joint angles across 20 degrees of freedom, designed for upper limb amputees. Our offline analysis shows 0.8 correlation between predicted and actual hand movements. The system functions as an integrated pipeline with three key components: (1) a VR-based data collection platform, (2) a transformer-based model for EMG-to-motion transformation, and (3) a real-time calibration and feedback module called ALVI Interface. Using eight sEMG sensors and a VR training environment, users can control their virtual hand down to finger joint movement precision, as demonstrated in our video:  \href{https://youtu.be/Dx_6Id2clZ0?si=tpt7wkDDqvQ54EDq}{youtube link}.

\end{abstract}

\section{Introduction}

Upper limb amputation has substantial physical, psychological, and occupational impacts on individuals \cite{shahsavari2020}. Even the most advanced bioelectric prostheses cannot completely solve the problem of low degree of freedom and control flexibility, which is relevant for the users. The main challenge is to create a universal and convenient control system for the prosthesis, simulating the natural control of a real hand. 
In recent works \cite{liu2021neuropose, sussillo2024generic}, authors have presented systems that convert muscle electrical signals from the surface of the forearm (sEMG) into precise hand movements in healthy people. However, for people with amputation, creating such a system remains difficult due to the absence of target movements to train the decoders.\cite{farina2014, cordella2016}
In this study, we present a system for decoding individual finger movements using sEMG signals for people with hand amputation in real-time. Our approach includes:
\begin{itemize}
    \item a VR setup for collecting paired datasets of finger movements and forearm muscle activity in amputees, which provides essential training data;
    \item a transformer-based model for decoding individual finger movements from sEMG signals, which processes this data;
    \item the ALVI Interface, a real-time system that enables adaptive control and visualization of a virtual hand with 20 degrees of freedom.
\end{itemize}

\section{Methods}

\subsection{Dataset}
\begin{figure}[h]
    \begin{center}
    \includegraphics[width=0.9\linewidth]{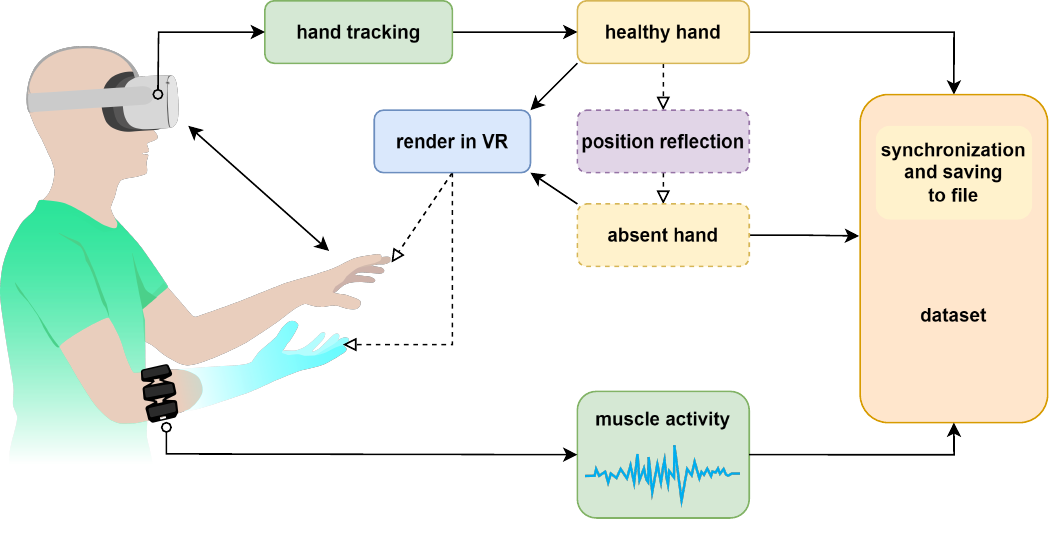}
    \end{center}
    \caption{\textbf{Data collection system.} Participant with VR headset and sEMG armband performs movements with intact hand. System tracks movements, mirrors them to create virtual model of absent hand, and records muscle activity from residual limb.}\label{fig:diagram_data_collection}
\end{figure}

We developed a VR application to collect accurate hand movement data from amputees, crucial for training our sEMG-based decoding algorithm \cite{miranda}. The system consists of three main components: an experimental environment, a hand reflection module, and a data aggregation module (figure \ref{fig:diagram_data_collection}). The VR environment features a display screen that shows instructions and guides users through specific movement sequences. This controlled setting enables participants to practice diverse hand motions while receiving real-time visual feedback. The hand reflection module was specifically designed to precisely capture target finger positions for the amputated hand. This is accomplished by tracking the coordinates of the fingers on the participant's intact hand using the Oculus Quest hand tracking system. These coordinates are then mirrored to create a virtual 3D model of the absent hand, reflecting the movements of the intact hand in a symmetrical manner. The data aggregation module synchronizes all input data, including finger positions and signals from sEMG sensors (obtained via wireless Myo Armband by Thalmic Labs). We used the open-source Lab Streaming Layer (LSL) framework \cite{kothe2014} to facilitate precise time-synchronized streaming of all data channels in real-time.

We implemented real-time inference for prosthetic hand control. After a brief calibration with simple gestures, users can perform any movement and see the virtual hand respond in the real-time.

\paragraph{Experiment}

Our experiment included 72 daily-life gestures (45 dynamic, 27 static), performed symmetrically with both hands. We had 22 participants (20 non-amputees, 2 amputees), each completing 3 sessions on different days: 2 for training and 1 for testing. Sessions lasted 1-1.5 hours, with movements repeated for 1 minute each.

\paragraph{Data preprocessing}

Raw EMG activity (200 Hz sampling frequency) is normalized to the [-1, 1] range using min-max scaling. Target movements are encoded as quaternions for 21 joint orientations, normalized relative to the palm position. We extract 4 angles for each finger, following the approach presented in NeuroPose3D. The electrode order for the left-hand data is rearranged to match the right-hand configuration to enable cross-hand compatibility. Movement data is downsampled from 40 Hz to 25 Hz for real-time processing. Our many-to-many approach uses a 1.28-second input window (8 channels, 256 time points) to predict 32 time points of 20 movement-encoding variables.

\subsection{Model}

We introduce HandFormer, a transformer-based architecture designed for EMG-to-motion translation. The model consists of an Encoder and a Decoder, optimized for processing EMG signals and generating hand movements. The EMG data are split into patches, each patch representing one electrode's activity over 8 time points (input shape: [8, 256], patch size (1, 8), resulting in 256 total tokens).
The encoder processes tokenized EMG data to extract relevant features. The Decoder employs a Perceiver-like architecture \cite{jaegle2021perceiver} with 32 learnable queries that match the number of movement frames. Notably, we use non-autoregressive prediction, which empirically outperforms autoregressive approaches for our task.
HandFormer's pretraining consists of two sequential stages. The first stage, EMG Encoder Pretraining, implements a masked autoencoder architecture \cite{he2022} with 70\% token masking to learn EMG signal patterns. In the second stage, we use the pretrained encoder weights for full model training, optimizing hand pose predictions using L1 loss between the predicted and target joint angles.

\begin{figure}[h]
    \centering
    \includegraphics[width=0.8\linewidth]{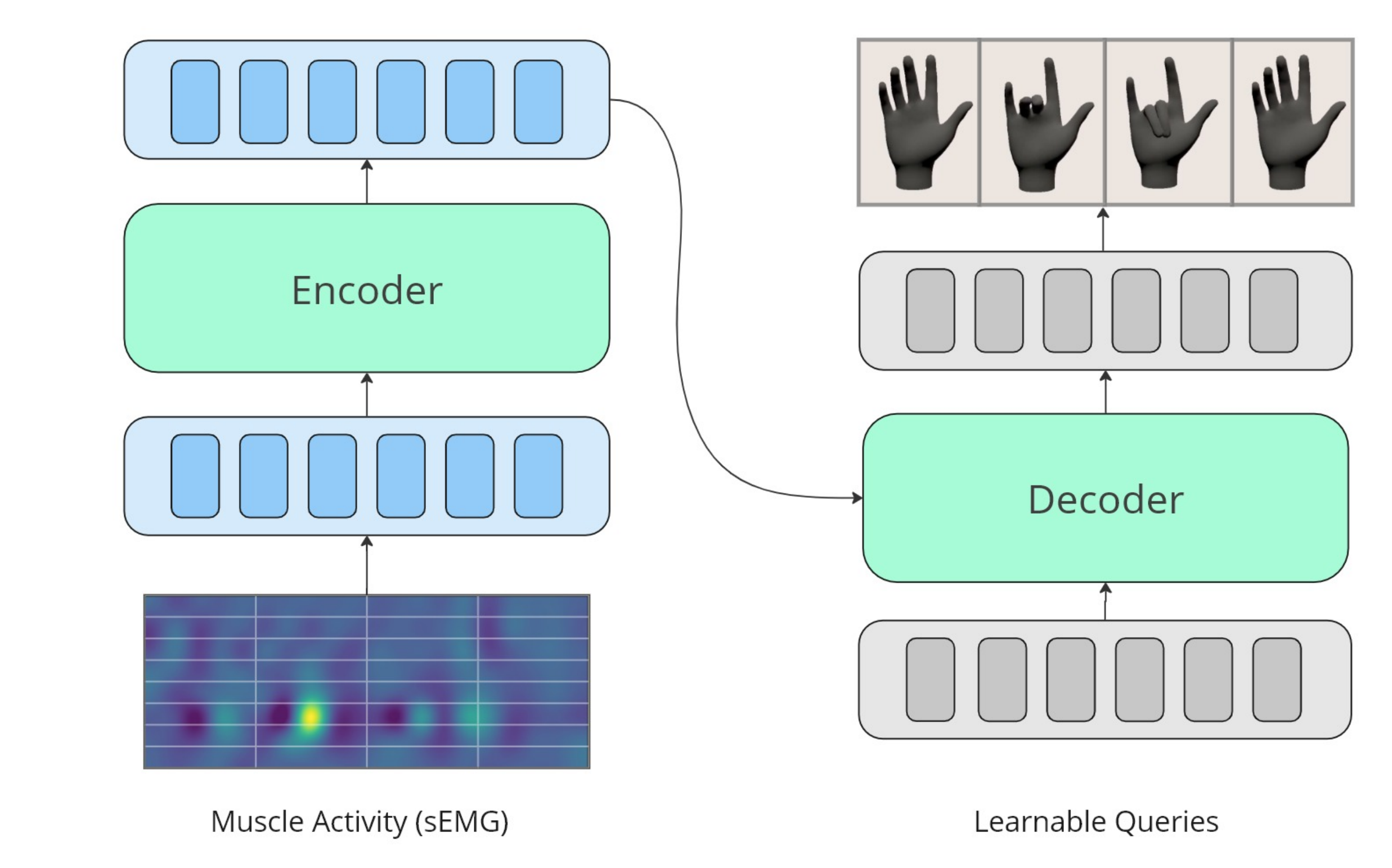}
    \caption{\textbf{Architecture of the model.} The HandFormer architecture transforms muscle activity (sEMG) into hand movements through a two-stage process. The Encoder (left) tokenizes sEMG signals from 8 channels into patches and extracts relevant features. The Decoder (right) employs a Perceiver-like architecture with 32 learnable queries corresponding to predicted movement frames. This non-autoregressive design enables efficient real-time translation of muscle signals into precise finger joint angles across 20 degrees of freedom.}
    \label{fig:model}
\end{figure}

\subsection{ALVI Interface}
\begin{figure}[h]
    \centering
    \includegraphics[width=\linewidth]{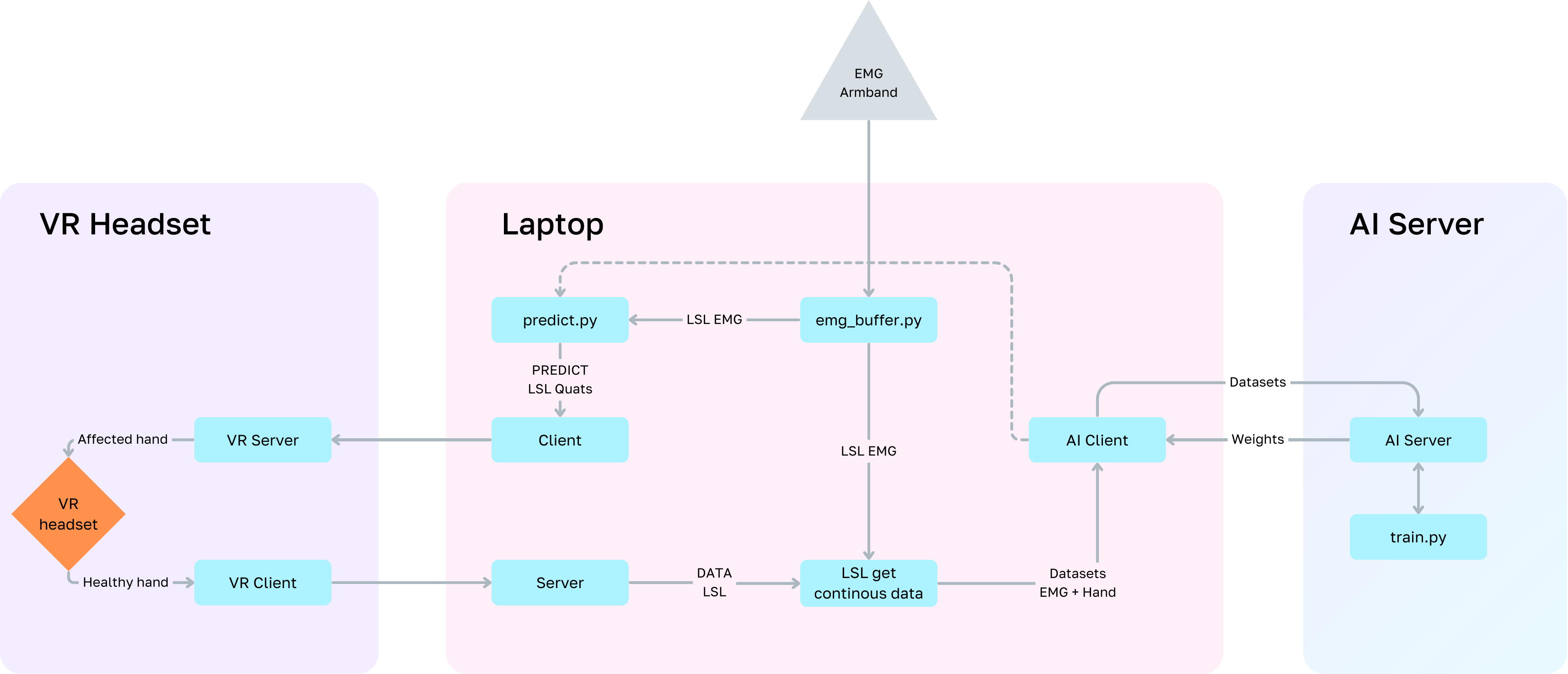}
    \caption{\textbf{ALVI Interface system architecture.}.}
    \label{fig:vr_app}
\end{figure}

Building upon our data collection platform, ALVI Interface extends the system from passive recording to active bidirectional control. While the initial system focuses on capturing training data, ALVI adds real-time decoding, visual feedback, and adaptive tuning capabilities, enabling users to immediately see their intended movements and participate in refining the model through interactive training. This evolution creates a comprehensive motion decoding system that enables practical, personalized control for amputees. (Figure \ref{fig:vr_app}).

\textbf{System Architecture:}
The distributed pipeline connects four key components: sEMG armband (input), VR headset (visualization), laptop (processing), and AI server (adaptation). Laptop processes sEMG signals to generate real-time hand movement predictions displayed in VR. Simultaneously, data is streamed to the AI server, which continuously updates the prediction model and sends improved weights back to the laptop, creating an adaptive learning loop between the system and user.

\textbf{Interactive Adaptation:}
ALVI Interface implements a novel approach to model calibration through interactive real-time training. During a 10-minute session, users perform movements while simultaneously observing their virtual hand's response, allowing them to:
\begin{itemize}
    \item Immediately see the quality of movement reconstruction
    \item Focus on specific gestures that need improvement
    \item Actively guide the training process based on visual feedback
\end{itemize}

The system continuously finetunes the pretrained HandFormer model to the user's sEMG patterns, updating weights every 10 seconds. This interactive loop combines both new and historical data, ensuring that each session builds upon previous ones while adapting to current conditions. The challenge of implementing closed-loop adaptive systems with real-time constraints is also being addressed in brain-computer interfaces, as demonstrated by the BRAND platform \cite{brand}.

\textbf{Real-time Performance:}
The system processes sEMG signals through a 256-point sliding window at 200 Hz, predicting 32 frames of movement at 25 Hz. It uses the most recent predicted frame for immediate control, with exponential moving average applied to joint angles for smoothness. This enables responsive, natural hand control with minimal latency.

After initial adaptation, the Interface runs independently on the laptop, enabling seamless hand control in VR without requiring constant server connection. This architecture provides a practical foundation for long-term prosthetic control in virtual environments, significantly enhancing user autonomy and experience.




\section{Results}

Our system demonstrated strong performance in both quantitative metrics and qualitative assessments. The evaluation was conducted across three key dimensions: offline accuracy, real-time performance, and user experience.
\paragraph{Quantitative Performance}
In offline testing with 22 participants (20 non-amputees, 2 amputees), the system achieved finger movement reconstruction with a correlation of 0.86 for non-amputee participants and 0.80 for amputee participants. The mean angular error was 8.09° and 14.50° respectively. Notably, our system is among the first to demonstrate such high performance levels for amputee users.

\begin{table}[htbp]
    \centering
    \caption{Performance comparison between non-amputees and amputees}
    \label{tab:performance_comparison}
    \begin{tabular}{
        l
        S[table-format=2.2]
        S[table-format=1.2]
    }
    \toprule
    \textbf{Subject Group} & 
    {\textbf{Angle error (\si{\degree})}} & 
    {\textbf{Correlation}} \\
    \midrule
    Non-amputees & 8.09 & 0.86 \\
    Amputees     & 14.50 & 0.80 \\
    \bottomrule
    \end{tabular}
\end{table}

\paragraph{Real-time Performance}
The system operates in real-time at 25 Hz with a latency of 51.2 ms (1.28-second window / 25 frames). After a 10-minute calibration period, users can perform natural hand movements in VR with smooth response. The continuous adaptation mechanism maintains consistent performance throughout extended usage sessions, with model weights updating every 10 seconds based on new sEMG patterns.

\paragraph{User Experience}
We conducted extensive qualitative evaluation with amputee participants (n=2) over multiple sessions. Our participants' interaction with the system revealed an interesting learning dynamic. Initially, users performed predefined gestures without system feedback to establish baseline data collection. After implementing real-time feedback through ALVI Interface, we observed significant improvements in control quality. Within the first 10 minutes of interactive training, users gained precise control over individual finger movements, showing rapid adaptation to the system.

A particularly interesting finding emerged after 30 minutes of use, when participants reported a mutual learning phenomenon - both the system and users adapted their behavior to achieve better results, similar to findings reported in other myoelectric interfaces \cite{hahne2017}.

Users noticed they were unconsciously adjusting their muscle activation patterns to match the model's expected inputs, while the system continuously refined its predictions based on user behavior.

The system's ability to retain personalized models across sessions proved valuable. Each subsequent session required less adaptation time, as both the system and user retained their learned patterns from previous interactions. This continuous improvement cycle created an increasingly natural and responsive interface, with users reporting more intuitive control in each session.

These observations suggest that our approach of combining initial structured training with interactive learning leads to a more personalized and effective user experience. The decreasing adaptation time across sessions indicates successful long-term learning on both the system and user sides.

\section{Discussion}





Our findings indicate that ALVI Interface can provide high-fidelity finger movement decoding in real-time using sEMG signals, even for individuals with upper limb amputation. A key challenge remains the inherent variability of sEMG signals—affected by electrode placement, muscle fatigue, and day-to-day fluctuations—necessitating regular calibration. Our co-adaptive approach offers a practical solution: the system continuously updates the model based on user input while users adapt their muscle activation patterns, resulting in rapid proficiency gains within each session.

This co-adaptation capability is particularly valuable for individuals with amputation, as they can quickly learn fine motor control in VR. Over multiple sessions, participants reported a growing sense of intuitive control, suggesting that sustained use further refines both the user’s activation strategy and the system’s decoding performance. Beyond prosthetic control, the same methodology could benefit stroke or injury rehabilitation by providing real-time visualization of intended movements, enhancing both patient motivation and clinical insights.

While the results are promising, larger clinical trials with more amputees are needed to generalize these findings and address known challenges in myoelectric prosthesis control \cite{chadwell2016}. Future work will focus on improving the long-term stability of the interface, integrating position-invariant sEMG decoding, and exploring advanced adaptation techniques to reduce calibration frequency. Further research should also investigate integrating ALVI with physical prostheses for real-world tasks, as well as exploring how VR-based training might accelerate users’ functional recovery.

\bibliographystyle{unsrt}
\bibliography{alvi_interface}

\end{document}